\documentclass[runningheads]{llncs}

\usepackage{eccv}

\usepackage{eccvabbrv}

\usepackage{graphicx}
\usepackage{booktabs}
\usepackage[export]{adjustbox}
\usepackage[accsupp]{axessibility}  
\usepackage{multirow}
\usepackage{tabularx}

\usepackage{algorithm}
\usepackage{algorithmic}
\usepackage{amsmath}

\usepackage{hyperref}

\usepackage{orcidlink}

\begin{document}

\renewcommand{\topfraction}{0.95}
\renewcommand{\bottomfraction}{0.95}
\renewcommand{\textfraction}{0.02}
\renewcommand{\floatpagefraction}{0.95}

\title{Revisiting Parameter Redundancy in Vision-Language-Action Models: Insights from VLM-to-VLA Adaptation}

\titlerunning{Revisiting Parameter Redundancy in VLA}



\author{
Fengnian Zhang\inst{1,2}$^{*}$ \and
Tao Huang\inst{3}$^{*}$ \and
Siyu Xu\inst{4} \and
Zhong Jin\inst{1}$^{\dagger}$ \and
Chang Xu\inst{4}
}

\authorrunning{F. Zhang et al.}

\institute{
Computer Network Information Center, Chinese Academy of Sciences\\
\email{fnzhang@cnic.cn, zjin@sccas.cn}
\and
University of Chinese Academy of Sciences
\and
School of Computer Science, Shanghai Jiao Tong University\\
\email{t.huang@sjtu.edu.cn}
\and
School of Computer Science, The University of Sydney\\
\email{\{s.xu,c.xu\}@sydney.edu.au}
}

\maketitle

\begingroup
\renewcommand\thefootnote{}
\footnotetext{$^{*}$ The authors contributed equally. 
$^{\dagger}$ Corresponding author.}
\endgroup

\begin{abstract}
Vision-Language-Action (VLA) models have made significant strides in embodied intelligence by integrating the powerful representations of pre-trained Vision-Language Models (VLMs). However, the massive parameter scale of VLAs imposes a heavy computational burden, and these models exhibit extreme sensitivity to parameter pruning. Current paradigms often treat the resulting performance degradation as inevitable, relying on fine-tuning or low-rank corrections to recover efficacy. We challenge this convention by questioning whether the removed parameters are truly redundant if VLA pruning necessitates performance recovery to be effective, or if this paradigm masks the indiscriminate pruning of critical parameters. We revisit parameter redundancy through the lens of VLM-to-VLA adaptation, first quantifying the spatial distribution of parameter divergence during adaptation to reveal structured patterns across different modules. Subsequently, we introduce controlled pruning as a diagnostic probe: by comparing the direct impact of removing different parameter subsets on VLA performance without any fine-tuning, we establish a causal link between adaptation-induced divergence signals and functional contributions. Based on the discovered modular heterogeneities, we design a multi-module joint pruning scheme. Evaluations on the LIBERO benchmark demonstrate that our approach reduces the parameters of OpenVLA and $\pi_{0.5}$ by 12\%--30\% while maintaining approximately 90\% of the original performance without any post-pruning recovery. In contrast, existing parameter pruning criteria result in total performance collapse when evaluated under the same recovery-free constraints. Our study reveals the parameter evolution mechanism in VLA adaptation and provides a new path for deploying efficient, robust robotic policies in resource-constrained environments. Code is available at \url{https://github.com/Niannnnnn/VLA_Parameter_Redundancy_VLM2VLA}.

\end{abstract}

\section{Introduction}
\label{sec:intro}

Vision-Language-Action (VLA) models have rapidly become the core paradigm in embodied intelligence. By adapting large-scale pre-trained Vision-Language Models (VLMs) to downstream action and control tasks, VLAs inherit rich cross-modal representations, showcasing significant generalization advantages in environment variations, object diversity, and task compositions \cite{sapkota2025vision,ma2024survey,kawaharazuka2025vision,xu2025affordance}. However, this ``inheritance'' introduces a long-overlooked yet critical question: \textbf{during the VLM-to-VLA adaptation, which parameters truly participate in action generation, and which are merely structural residues from pre-training?}

This issue is amplified by the sheer scale of modern VLA architectures, which typically retain the vast majority of VLM parameters, leading to substantial computational and storage overhead. Various efficiency optimization methods have been proposed \cite{yu2025survey,guan2025efficient}, including parameter pruning, token pruning, layer pruning, quantization, and lightweight architectural design. However, a recurring phenomenon is that VLA models are extremely sensitive to parameter removal; even moderate pruning can lead to a ``performance cliff'' where task success rates plummet and model behavior becomes nearly non-functional.

Facing this sensitivity, current mainstream strategies are pragmatic and focused on engineering fixes \cite{chen2025rlrc,jabbour2025don}: they treat post-pruning performance degradation as an unavoidable side effect and compensate for it through post-hoc recovery means such as fine-tuning or low-rank adaptation. A pruning method is deemed ``successful'' once the model performance reaches an acceptable level during the recovery phase. We fundamentally question this premise. We argue that the widespread reliance on performance recovery masks a more critical issue: \textbf{if a pruned VLA model must re-learn to recover its functionality, were the removed parameters truly redundant?} 

This observation prompts us to shift the focus from traditional inquiries like ``how to prune more'' or ``how to achieve better recovery'' to a more foundational question: \textbf{when pruning necessitates explicit performance recovery, does it reflect true redundancy or the indiscriminate pruning of vital parameters?} This question remains under-discussed because existing methods often conflate two distinct problems: the identification of redundant parameters and the repair of pruning-induced damage. Our position is that parameter redundancy in VLAs is not equivalent to that in traditional CNNs or LLMs. Unlike models trained end-to-end on a single task distribution, VLAs are constructed through a structured adaptation process from general vision-language understanding to robotic control \cite{kim2024openvla,zitkovich2023rt,li2024cogact,mitra2025mechanistic}. During this process, parameters are selectively reused, re-weighted, or stabilized to support action-centric computation paths.

Based on these reflections, we revisit VLA parameter redundancy from the perspective of the VLM-to-VLA adaptation process. We first quantify the parameter divergence during this transition to reveal structured distribution patterns across different components. However, the relationship between parameter divergence and importance is not necessarily monotonic. Different modules may follow distinct adjustment mechanisms: in some modules, significantly changed parameters may directly participate in action-centered decisions; in others, stable parameters might provide crucial cross-modal alignment. 

To systematically investigate this, we introduce controlled pruning as a diagnostic probe. By evaluating the direct impact of removing parameters with different divergence characteristics (e.g., pruning parameters with the largest vs. smallest divergence) on policy performance, we establish a causal link between divergence signals and functional importance. This allows us to verify if VLM-to-VLA adaptation signals can effectively identify VLA parameter redundancy.
Specifically, we systematically test the following hypotheses:
\begin{itemize}
    \item \textbf{Hypothesis I}: If pruning in VLA models requires post-pruning performance recovery, the removed parameters are unlikely to be truly redundant.
    \item \textbf{Hypothesis II}: The parameter differences introduced during VLM-to-VLA adaptation contain useful signals for identifying redundant parameters.
    \item \textbf{Hypothesis III}: The usefulness of VLM-VLA parameter differences for redundancy varies across modules.
    \item \textbf{Hypothesis IV}: When properly utilized, VLM-VLA parameter differences allow structured pruning without requiring post-pruning performance recovery.
\end{itemize}

Through systematic verification of these hypotheses, we establish a causal chain from adaptation signals to functional importance, and translate the resulting insights into a principled, recovery-free pruning scheme.
\noindent Our contributions are summarized as follows:
\begin{enumerate}
    \item We revisit VLA parameter redundancy through the lens of VLM-to-VLA adaptation, introducing a novel diagnostic framework that treats VLA  pruning as a controlled intervention to causally link adaptation-induced divergence with parameter functional importance.
    \item We reveal that the prevailing recovery-dependent paradigm conceals the indiscriminate removal of vital parameters, and show that the $\Delta W$ signals exhibit strong module heterogeneity, necessitating differentiated pruning strategies across vision encoders, language backbones, and cross-modal projectors in VLA models.
    \item We propose a multi-module joint pruning scheme that reduces VLA parameter scale and memory usage by 12\%--30\% while maintaining approximately 90\% of the original performance of OpenVLA and $\pi_{0.5}$ \textit{without any post-pruning recovery}, validating the high utility of VLM-to-VLA adaptation signals.
\end{enumerate}

\section{Related Work}
\label{sec:related_work}

\subsection{Vision-Language-Action (VLA) Models}
Recent VLA models have substantially improved embodied agents by adapting large-scale Vision-Language Models (VLMs) to robotic action generation. Pioneering works such as RT-2 \cite{zitkovich2023rt} and PaLM-E \cite{driess2023palm} have demonstrated zero-shot transfer potential across various tasks and environments. OpenVLA \cite{kim2024openvla} established a high-performance baseline for the open-source community by instruction-tuning the Prismatic VLM \cite{karamcheti2024prismatic}, while $\pi_{0.5}$ \cite{intelligence2025pi05visionlanguageactionmodelopenworld} adopted the modular and lightweight PaLI-Gemma \cite{beyer2024paligemma} architecture to balance inference efficiency and performance. Furthermore, works like CogACT \cite{li2024cogact} and Gr00t \cite{bjorck2025gr00t} explored the deep coupling between multimodal representations and action decision-making. Despite these advances, most VLAs inherit the massive parameter scale of their VLM backbones, while the functional roles of these parameters remain largely unexplored.

\subsection{Efficiency and Pruning in VLA Models}
To alleviate the computational burden of large-scale VLAs, researchers have proposed various optimization strategies, including parameter pruning \cite{chen2025rlrc,jabbour2025don,xu2026qvla}, token pruning \cite{xu2025vla,pei2025action,tan2025think,li2025sp,yang2025efficientvla}, layer pruning \cite{yang2025efficientvla,zhang2025mole,shukor2025smolvla,yue2024deer}, quantization \cite{wang2025bitvla,fang2025sqap,park2025saliency}, and lightweight architectural design \cite{liu2024robomamba,wen2025tinyvla}. Orthogonally, adaptive test-time compute allocation has also been explored to improve VLA efficiency \cite{li2026vla}. Among these, parameter pruning has most directly exposed the extreme sensitivity of VLA models.

RLRC \cite{chen2025rlrc} introduced structured pruning based on LLM-Pruner and Taylor importance criteria, utilizing fine-tuning and reinforcement learning (RL) to recover performance. However, empirical findings showed that without Supervised Fine-Tuning (SFT), performance could hardly be recovered even after 2 million RL steps, revealing the indispensability of post-hoc recovery in existing workflows. To address the VLA performance collapse caused by parameter pruning, GLUESTICK \cite{jabbour2025don} utilized SVD to extract principal directions of weight differences and added lightweight corrective terms during inference. While avoiding fine-tuning, it essentially remains a post-hoc remedy. 

Unlike studies focused on ``how to recover'', we challenge the premise that performance loss is inevitable. We argue that the heavy reliance on recovery reflects a failure to identify redundancy criteria. We explore a precision identification method based on adaptation features to enable direct pruning without any post-hoc compensation.

\subsection{Mechanistic Connections between VLM and VLA}
The performance of a VLA model is deeply coupled with the pre-training quality of its VLM backbone, yet the relationship is not a simple linear mapping.

VLM4VLA \cite{zhang2026vlm4vla} systematically compared various VLM backbones and found that VLM performance on standard benchmarks does not necessarily correlate with downstream VLA task performance. This suggests complex structural reorganization during the transfer from vision-language understanding to robotic control. Actions as Language \cite{hancock2025actions} highlighted the "catastrophic forgetting" of VLM knowledge during VLA fine-tuning. By re-labeling robotic actions as image-text pairs, they maintained VLM capabilities while ensuring VLA performance.

Existing studies primarily focus on final performance comparisons. In contrast, we treat the VLM-to-VLA adaptation process itself as an observable structured signal. By investigating the link between adaptation-induced parameter divergence and policy performance, we seek to understand adaptation from a parametric dimension and utilize it as the core criterion for identifying redundancy.

\section{Analysis Framework}
\label{sec:framework}

This section presents an analysis framework designed to systematically investigate parameter redundancy in Vision-Language-Action (VLA) models. Unlike existing studies that treat pruning merely as a performance optimization technique, we reframe it as a \textit{diagnostic probe} to explore the functional roles of parameters adapted from pre-trained Vision-Language Models (VLMs) in robotic manipulation and control tasks. This shift in perspective allows us to clearly decouple two issues often conflated in prior research: the identification of redundant parameters and the post-pruning recovery of performance.

\subsection{Reframing VLA Parameter Pruning as an Analysis Problem}
Parameter pruning, a classical method for reducing computational overhead in large-scale neural networks, has been extensively studied. However, in the context of VLA models, a recurring phenomenon has been observed: even the removal of a moderate proportion of parameters can lead to a drastic decline in task success rates. Such performance losses typically necessitate large-scale fine-tuning or low-rank adaptation to recover.

While these recovery strategies are effective in engineering practice, they implicitly accept a fundamental premise: that performance degradation is an inevitable cost of parameter pruning. We argue that this assumption masks a more fundamental question about \textbf{what pruning actually removes}: vital parameters undergoing functional reorganization, or genuine structural residues. 

\noindent Let $f(\cdot; W)$ denote a VLA model with parameters $W$. Given a parameter pruning operator $\mathcal{P}(\cdot; M)$ defined by a binary mask $M \in \{0,1\}^{|W|}$, most pruning methods implicitly optimize the following objective:
\begin{equation}
    \max_{M} \ \mathcal{S}\big(f(\cdot; \mathcal{P}(W; M))\big) \quad \text{subject to} \quad \|M\|_0 \leq k,
\end{equation}
where $\mathcal{S}(\cdot)$ represents the task success rate and $k$ is the sparsity budget. 

However, once a fine-tuning process $\mathcal{T}(\cdot)$ is introduced post-pruning, the final evaluated model becomes $f\big(\cdot; \mathcal{T}(\mathcal{P}(W; M))\big)$. In this case, the model's performance no longer reflects the intrinsic functional capability supported solely by the retained parameters; instead, it reflects the model's capacity for re-learning and self-reconstruction under structural perturbation. From an analytical standpoint, this raises a fundamental concern: \textbf{if a pruning intervention must rely on explicit performance recovery to be effective, were the removed parameters truly redundant?} Consequently, we reframe VLA parameter pruning as an analytical problem rather than an optimization goal. In this framework, pruning (without subsequent performance recovery) is treated as a \textit{controlled intervention} used to reveal the direct impact of parameter removal on policy behavior. Under this paradigm, the necessity of performance recovery is no longer evidence of successful redundancy identification, but rather a signal of its failure.

\subsection{VLM--VLA Parameter Difference as a Structural Signal}
VLA models are typically not trained from scratch but are initialized from pre-trained VLMs and adapted using robotic manipulation data. Let $W^{\text{VLM}}$ denote the parameters of a pre-trained VLM, and $W^{\text{VLA}}$ denote the corresponding parameters after adaptation. In the shared VLM backbone subspace, we define the relative parameter divergence as:
\begin{equation}
    \Delta W_{\text{rel}} = \frac{\| W^{\text{VLA}} - W^{\text{VLM}} \|_2}{\| W^{\text{VLM}} \|_2}.
\end{equation}
We hypothesize that $\Delta W$ is not mere random noise but contains structured information regarding the functional reorganization of weights during adaptation. To verify this, we visualized the weight divergence distributions for two representative pairs: $\langle\text{Prismatic, OpenVLA}\rangle$ and $\langle\text{PaLI-Gemma, } \pi_{0.5}\rangle$.

\begin{figure}[htbp]
    \centering
    \begin{subfigure}[b]{0.48\textwidth}
        \centering
        \includegraphics[width=\textwidth]{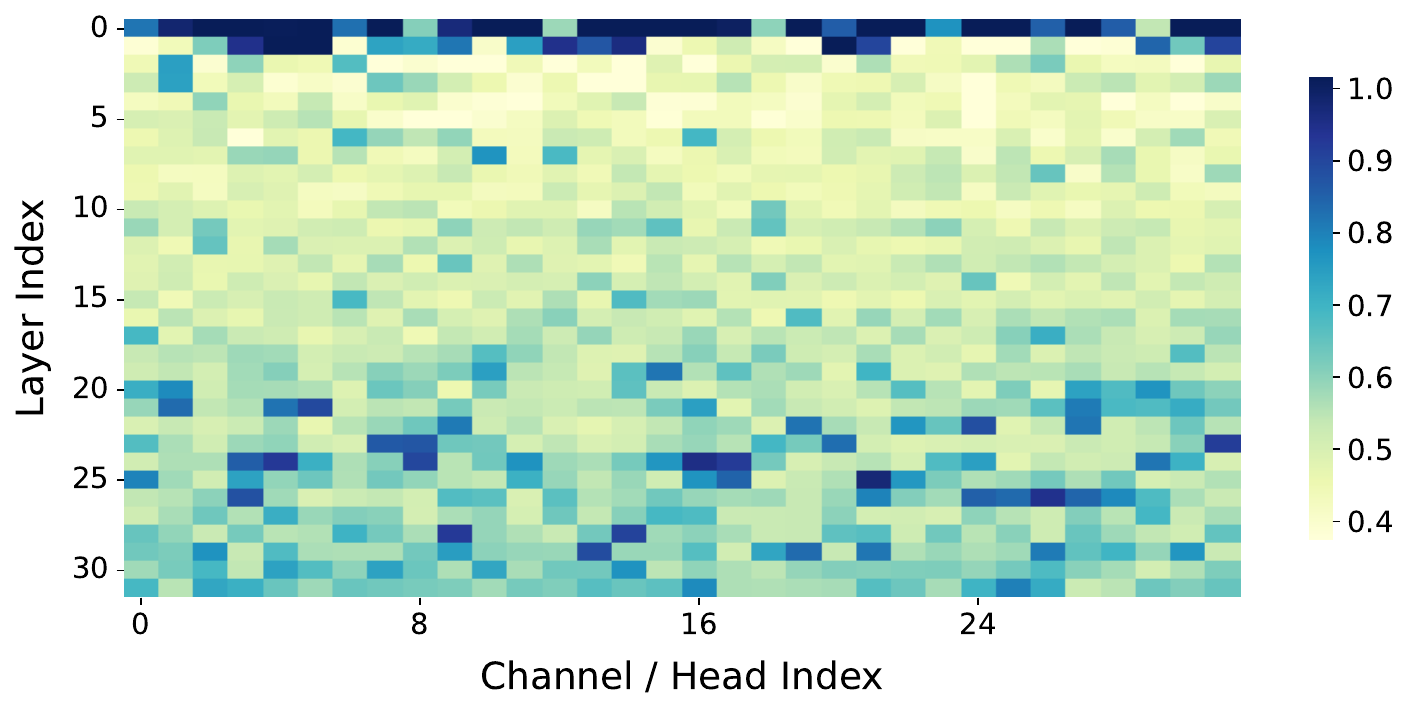}
        \caption{Llama 2 Attention Heads}
    \end{subfigure}
    \hfill
    \begin{subfigure}[b]{0.48\textwidth}
        \centering
        \includegraphics[width=\textwidth]{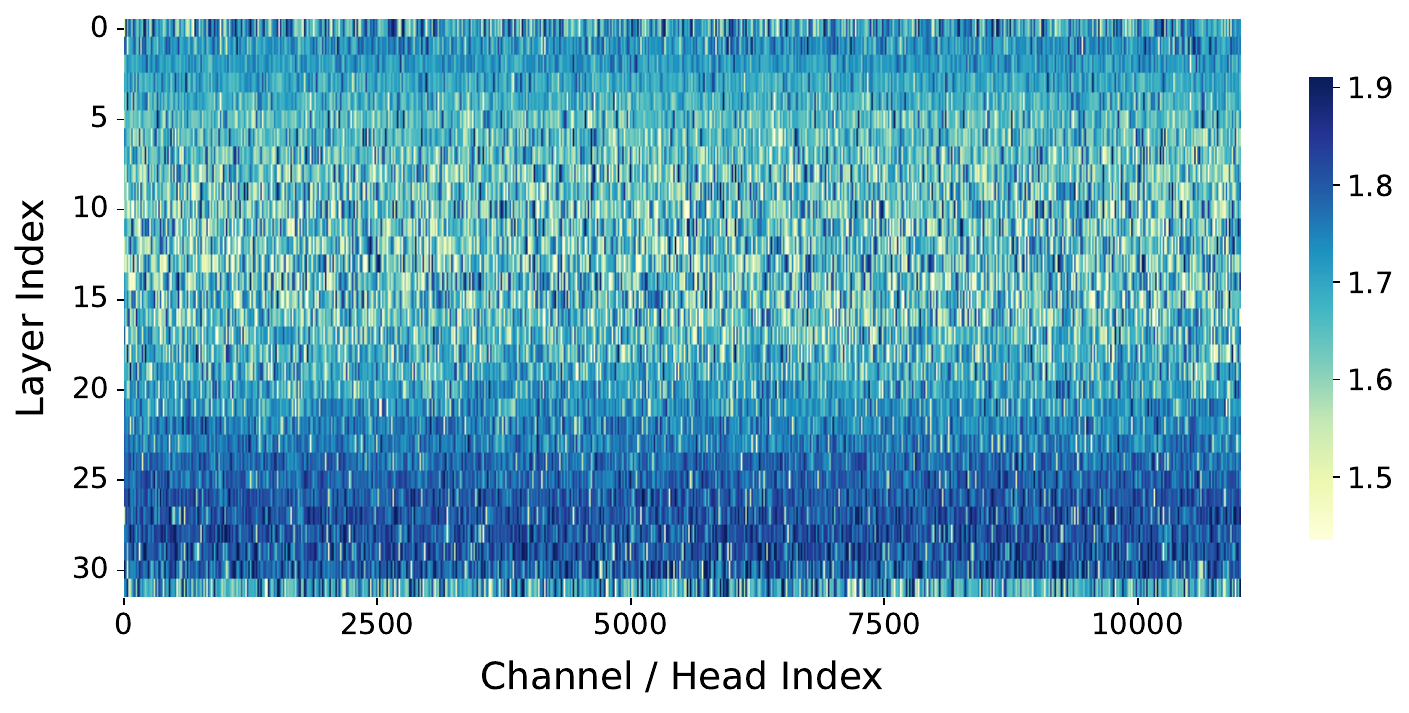}
        \caption{Llama 2 FFN Channels}
    \end{subfigure}

    \begin{subfigure}[b]{0.48\textwidth}
        \centering
        \includegraphics[width=\textwidth]{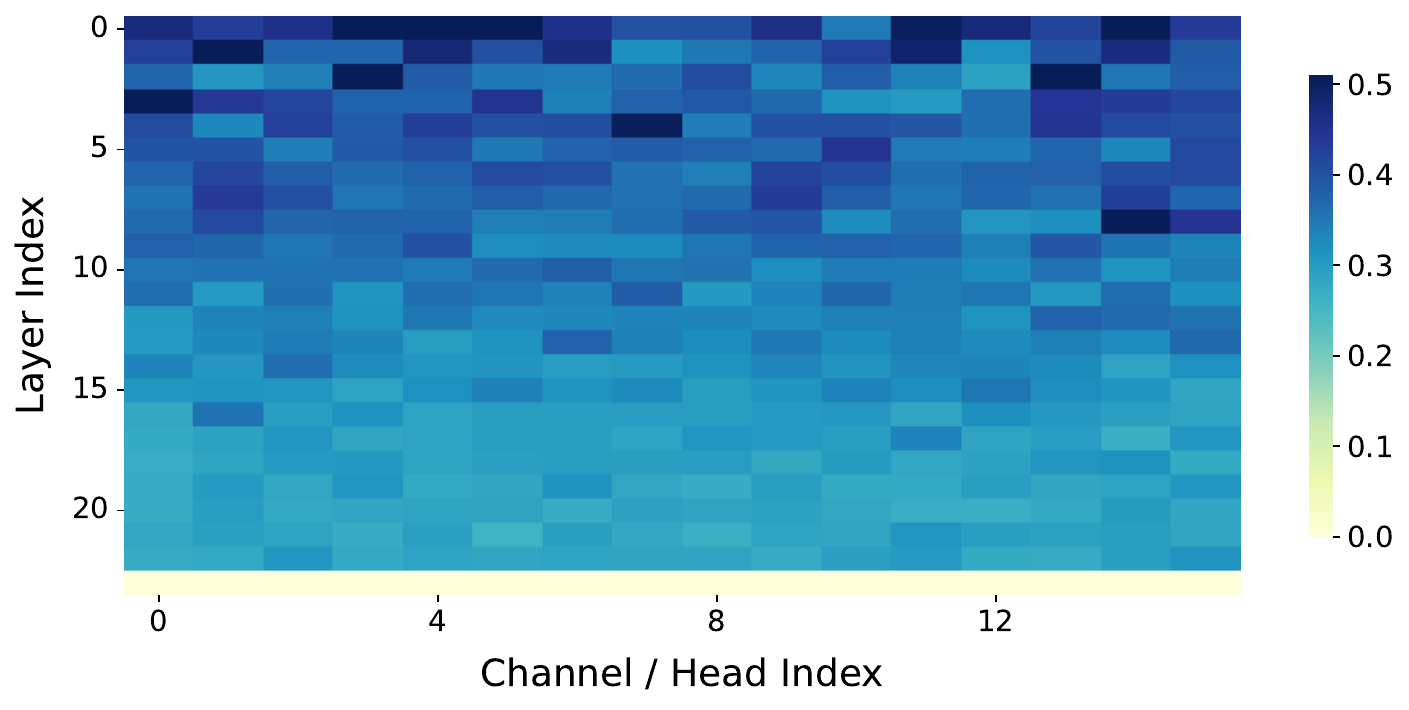}
        \caption{DINOv2 Attention Heads}
    \end{subfigure}
    \hfill
    \begin{subfigure}[b]{0.48\textwidth}
        \centering
        \includegraphics[width=\textwidth]{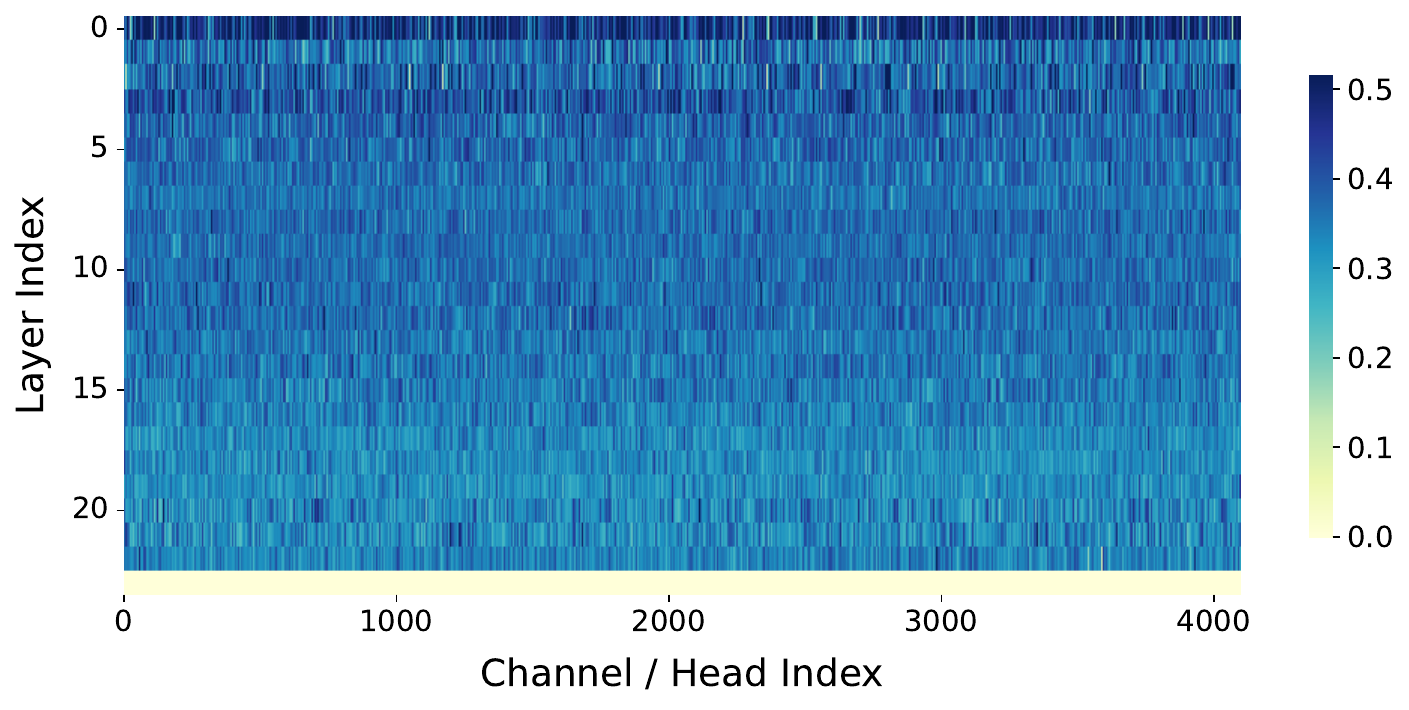}
        \caption{DINOv2 FFN Channels}
    \end{subfigure}

    \begin{subfigure}[b]{0.48\textwidth}
        \centering
        \includegraphics[width=\textwidth]{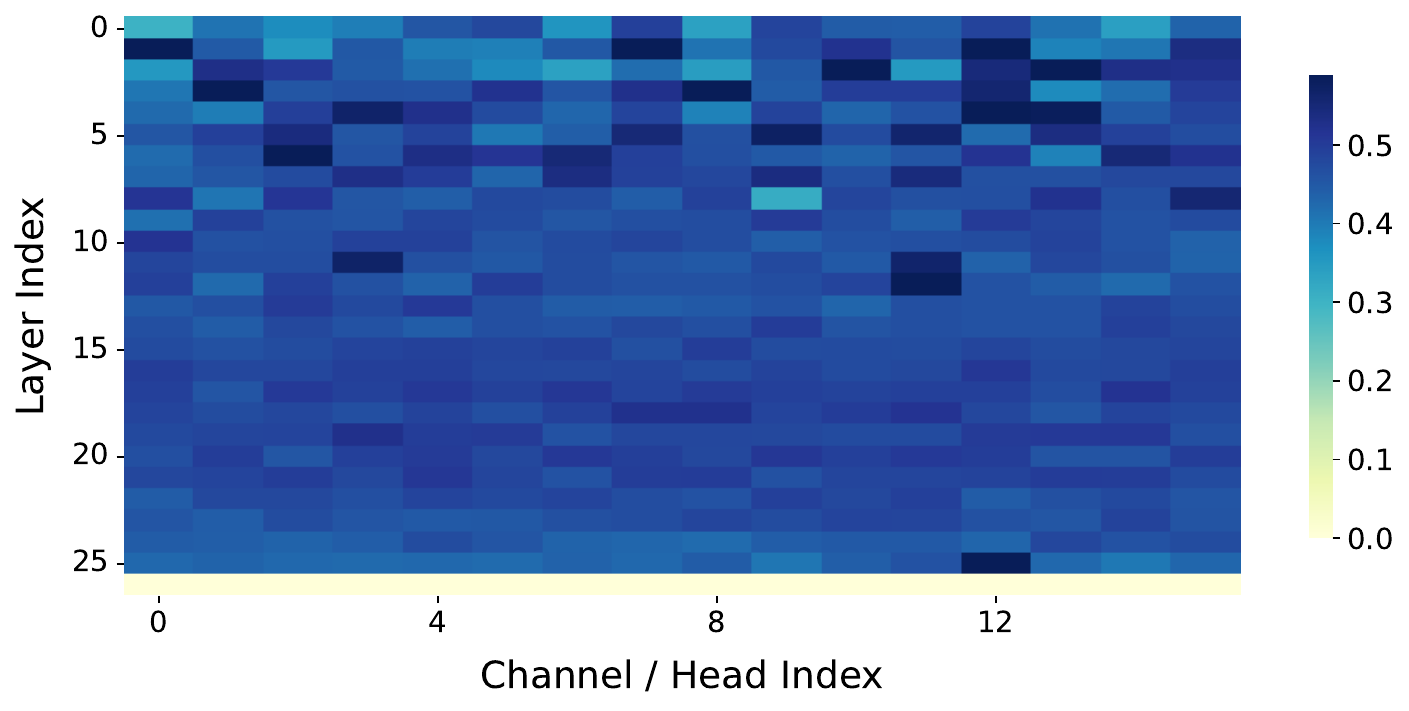}
        \caption{SigLIP Attention Heads}
    \end{subfigure}
    \hfill
    \begin{subfigure}[b]{0.48\textwidth}
        \centering
        \includegraphics[width=\textwidth]{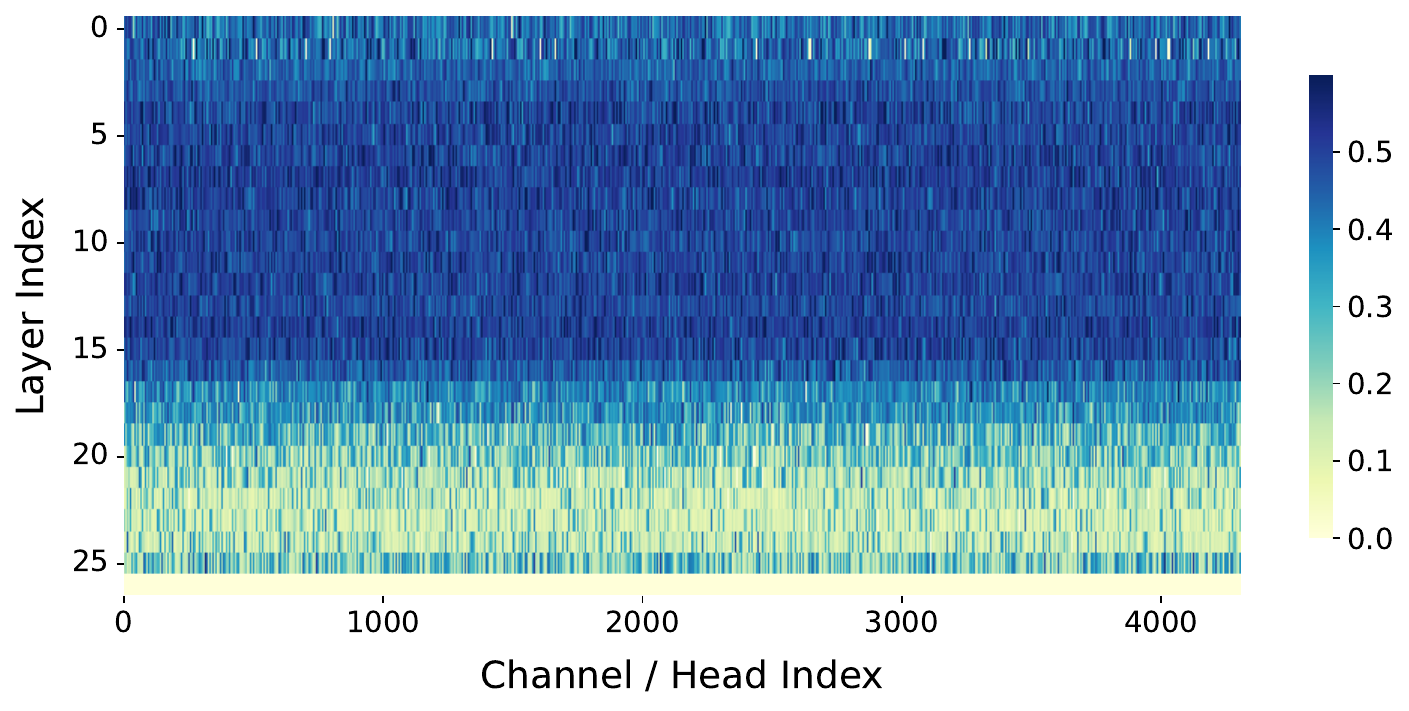}
        \caption{SigLIP FFN Channels}
    \end{subfigure}

    \begin{subfigure}[b]{0.6\textwidth}
        \centering
        \includegraphics[width=\textwidth]{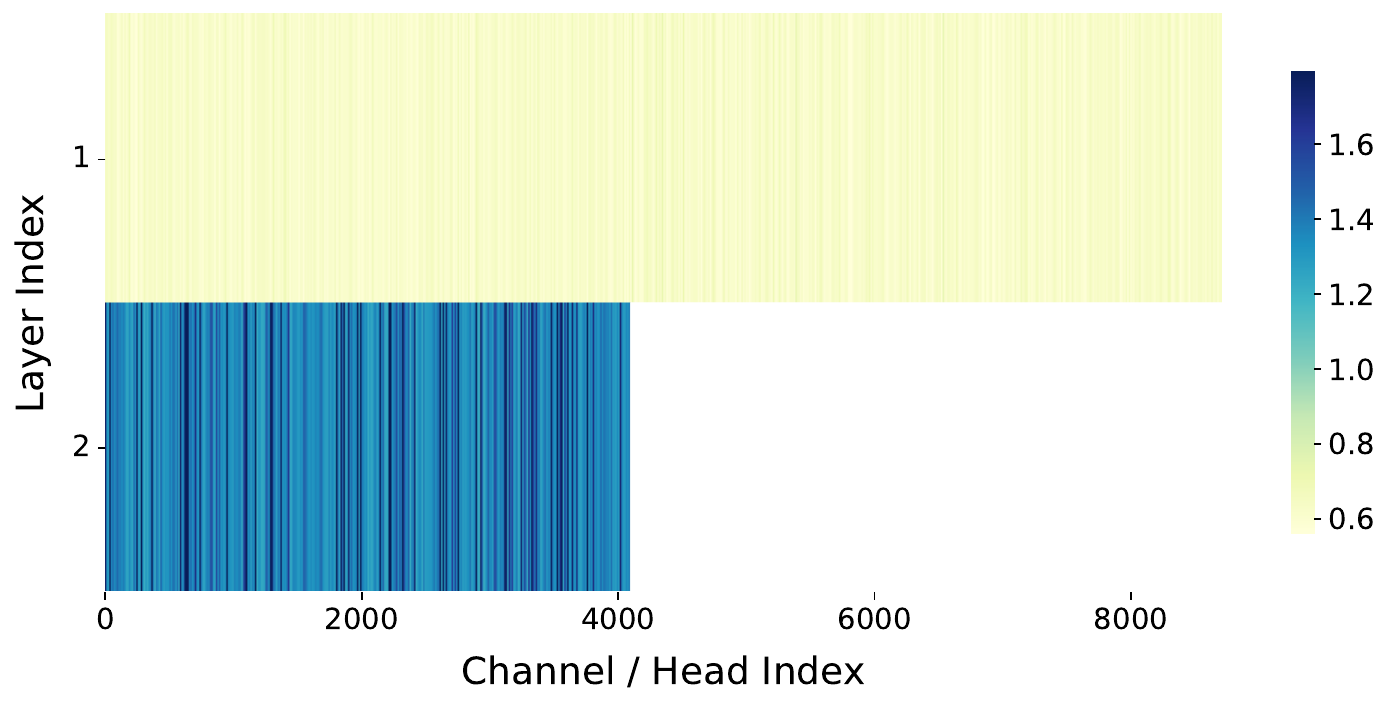}
        \caption{Projector FFN Channels}
    \end{subfigure}

    \caption{Visualizing the relative parameter divergence $\Delta W_{\text{rel}}$ between the \textbf{Prismatic} (VLM) and \textbf{OpenVLA} (VLA) model pair. 
    The color intensity indicates the magnitude of divergence: \textbf{darker blue} denotes significant parameter shifts, while \textbf{brighter yellow} represents minimal change. 
    Subfigures (a)--(g) display the divergence across different modules, calculated at the granularity of individual attention heads or FFN channels. 
    }
    \label{fig:openvla_heatmaps}
\end{figure}

In the \textbf{OpenVLA} (Prismatic-based) model, parameter divergence exhibits a highly non-uniform distribution across components (see Fig. 1):
\begin{itemize}
    \item \textbf{Language Model (Llama 2):} Parameter updates follow a distinct ``three-stage'' vertical distribution. The initial layers ($L_0$) undergo dense calibration to handle multimodal fusion after token injection; the middle layers ($L_1$--$L_{23}$) remain relatively stable; and the terminal layers ($L_{24}$--$L_{31}$) fluctuate again, reflecting the refined mapping to the action semantic space. In the Feed-Forward Network (FFN) modules, we observe inter-layer fluctuations and distinct ``strip-like'' sparsity in the channel dimension, suggesting that embodied knowledge is encoded in specific sub-channels rather than uniformly distributed.
    \item \textbf{Vision Backbone (DINOv2\&SigLIP):} For DINOv2, updates in Attention and FFN are concentrated in shallow layers, focusing on low-level visual cues like grasping and obstacle avoidance. Conversely, SigLIP shows stronger responses in deeper layers, with its FFN channel updates generally higher than those of DINOv2, serving as a semantic supplement to align visual features with the language model.
    \item \textbf{Projector:} Comprising FFN structures, its parameter changes show a monotonic increase. The late-stage mapping layers ($fc_2, fc_3$) fluctuate far more than the initial up-sampling layer ($fc_1$), indicating that semantic transformation layers near the language model entrance have higher update priority.
\end{itemize}

In the \textbf{$\pi_{0.5}$} (PaLI-Gemma-based) model, the modular design leads to more regular evolution patterns (see Fig. 2):
\begin{itemize}
    \item \textbf{Language Model (Gemma):} The Multi-Query Attention (MQA) mechanism shows high discriminative signals in head dimensions. Low layers ($L_0$) show minimal divergence, retaining general text-parsing priors; middle layers ($L_1$--$L_9$) exhibit large divergence, reflecting intense functional reorganization for redirection to embodied task attention; high layers ($L_{10}$--$L_{17}$) show a slight decrease, indicating stability after achieving the semantic-to-action mapping.
    \item \textbf{Vision Tower:} Attention modules show mild changes, suggesting well-preserved visual priors. FFN modules in the middle ($L_3$--$L_{15}$) and high ($L_{25}$--$L_{26}$) layers show strong divergence signals, providing a key basis for identifying redundancy in vision modules.
\end{itemize}

\begin{figure}[htbp]
    \centering
    \begin{subfigure}[b]{0.48\textwidth}
        \centering
        \includegraphics[width=\textwidth]{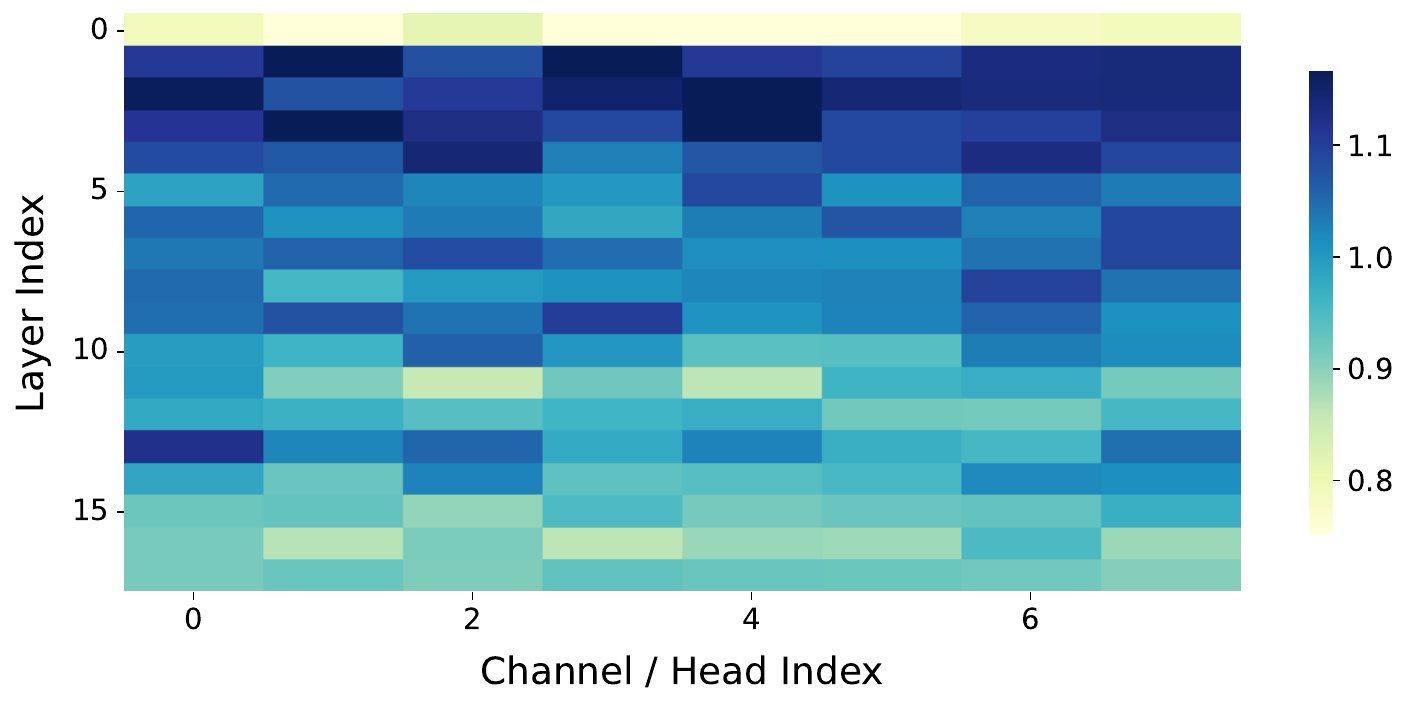} 
        \caption{Gemma Attention Heads}
    \end{subfigure}
    \hfill
    \begin{subfigure}[b]{0.48\textwidth}
        \centering
        \includegraphics[width=\textwidth]{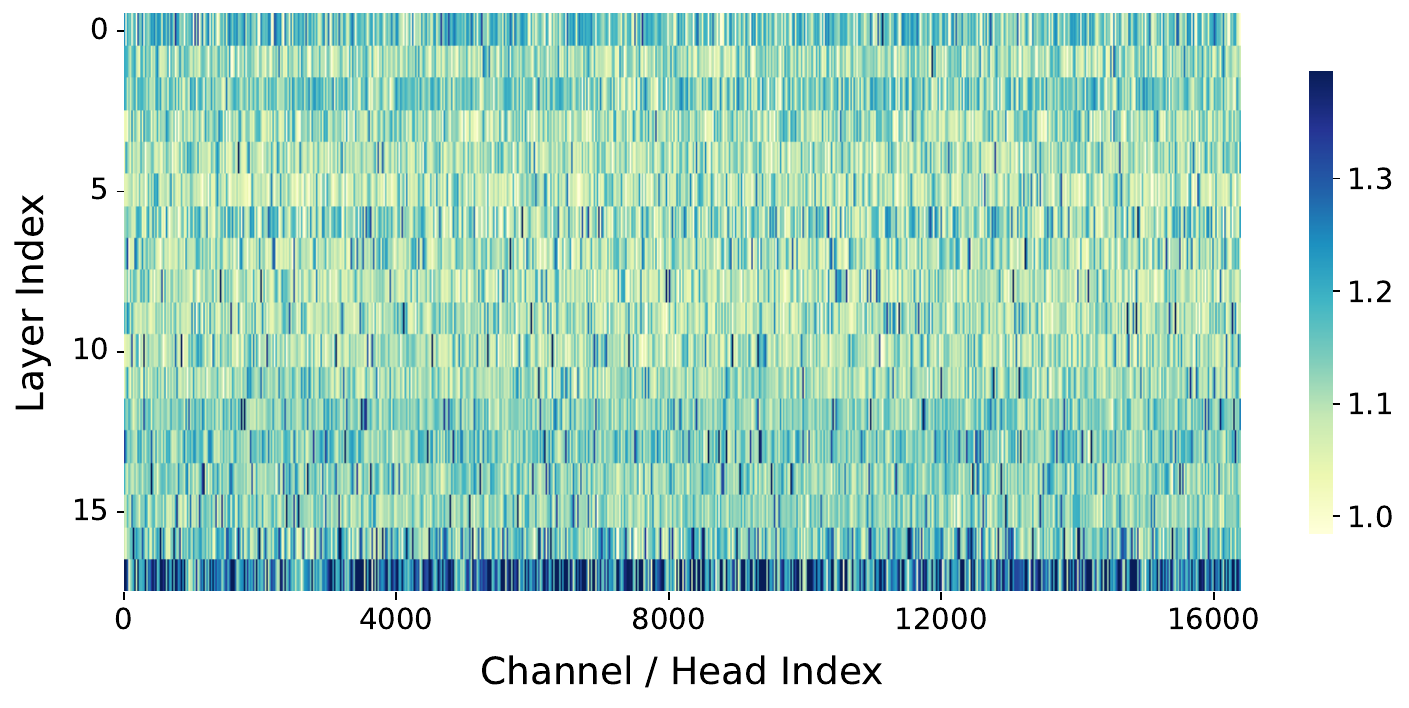}
        \caption{Gemma FFN Channels}
    \end{subfigure}

    \begin{subfigure}[b]{0.48\textwidth}
        \centering
        \includegraphics[width=\textwidth]{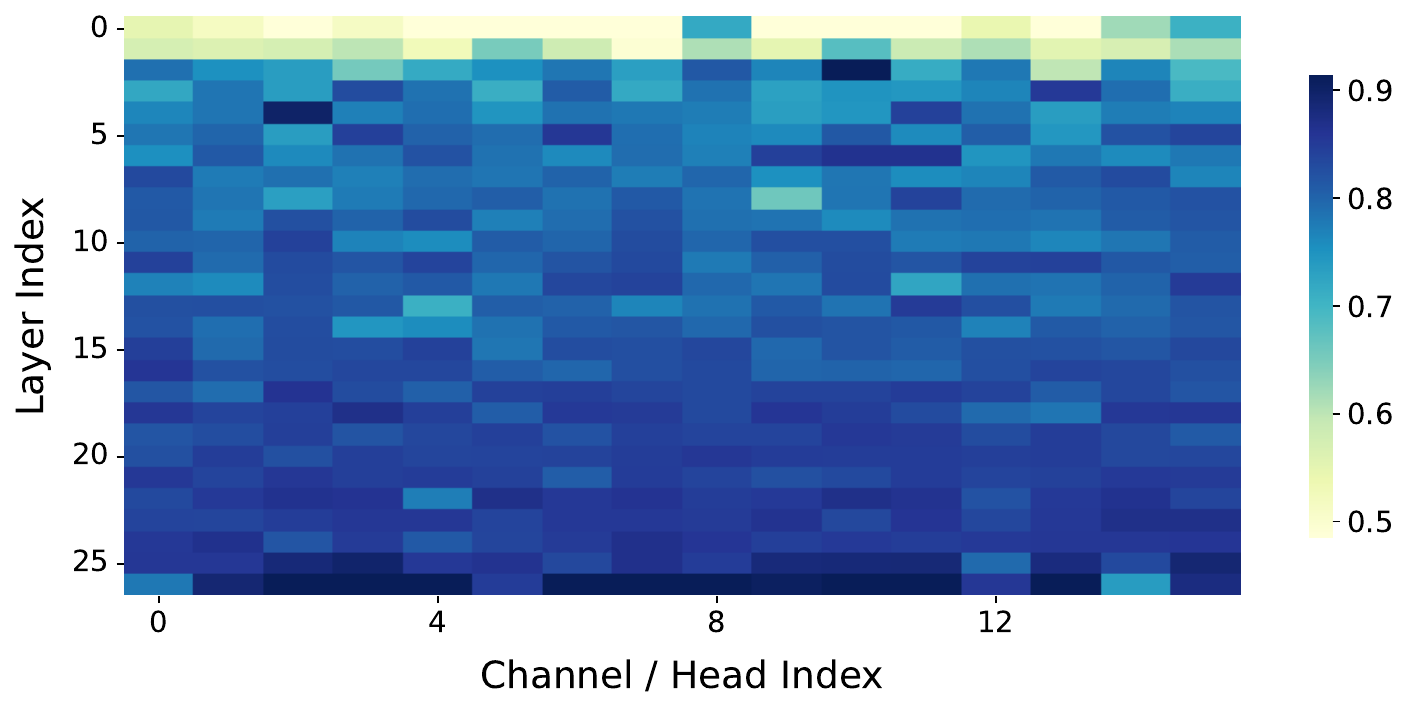}
        \caption{Vision Attention Heads}
    \end{subfigure}
    \hfill
    \begin{subfigure}[b]{0.48\textwidth}
        \centering
        \includegraphics[width=\textwidth]{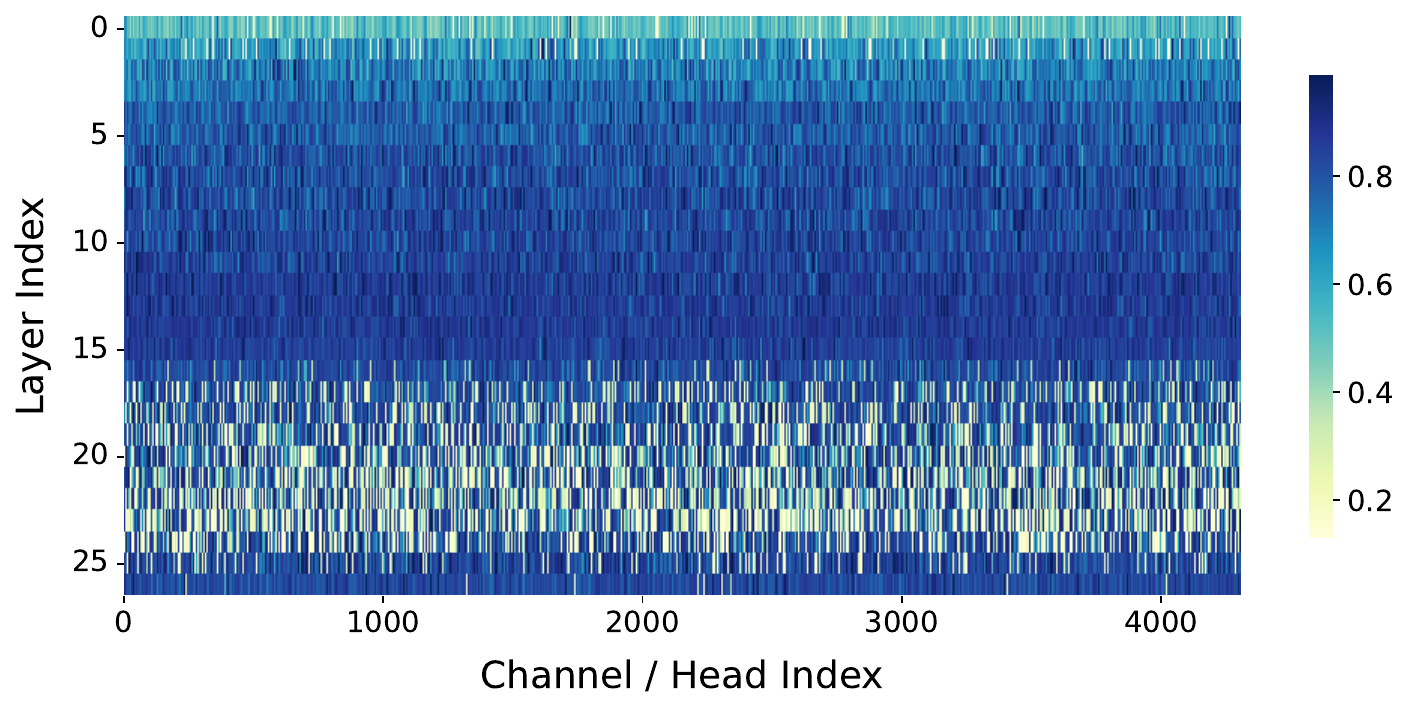}
        \caption{Vision FFN Channels}
    \end{subfigure}

    \caption{Visualizing the relative parameter divergence $\Delta W_{\text{rel}}$ for the \textbf{PaLI-Gemma} (VLM) and \textbf{$\pi_{0.5}$} (VLA) model pair. 
    The color intensity indicates the magnitude of divergence: \textbf{darker blue} denotes significant parameter shifts, while \textbf{brighter yellow} represents minimal change. 
    Subfigures (a)--(d) display the divergence across different modules, calculated at the granularity of individual attention heads or FFN channels.}
    \label{fig:pi05_heatmaps}
\end{figure}

These systematic observations lead to a crucial conclusion: parameter updates during VLM-to-VLA adaptation are not random noise but exhibit strong \textbf{structured heterogeneity} in both selective Attention-head reorganization and localized FFN-channel activation. This non-random distribution provides the empirical foundation for establishing precise VLA redundancy criteria and breaking the ``prune-then-collapse'' bottleneck.

\subsection{Controlled Pruning as a Diagnostic Instrument}
To link parameter divergence with model behavior, we introduce \textbf{controlled pruning} as a diagnostic tool. Given a target module $(g, h)$ and a pruning ratio $r$, we construct two complementary strategies based on $|\Delta W|$:
\begin{itemize}
    \item \textbf{Highest-difference pruning:} 
    \begin{equation}
        M^{\text{high}}(r) = \text{Top-}r\% \text{ parameters ranked by } |\Delta W|
    \end{equation}
    \item \textbf{Lowest-difference pruning:} 
    \begin{equation}
        M^{\text{low}}(r) = \text{Bottom-}r\% \text{ parameters ranked by } |\Delta W|
    \end{equation}
\end{itemize}

The pruned parameters are defined as $W^{\text{pruned}} = \mathcal{P}(W^{\text{VLA}}; M)$, where the pruning is applied directly \textit{without any fine-tuning}. To ensure the reproducibility of our diagnostic probe and clarify how $\Delta W$ is aggregated across multi-matrix modules (e.g., LLM FFNs), we provide the systematic mask construction process in Algorithm~\ref{alg:mask_gen}.

\begin{algorithm}[htbp]
\caption{Module-Aware VLA Mask Construction}
\label{alg:mask_gen}
\begin{algorithmic}[1]
\REQUIRE $W^{\text{VLM}}, W^{\text{VLA}}$, Pruning ratio $r$, Strategy $\in \{\text{Lowest, Highest}\}$
\ENSURE Binary masks $M$ for VLA parameters
\STATE \textbf{Initialize:} Importance score pool $S \leftarrow \emptyset$
\FOR{each module $\mathcal{L} \in$ \{LLM, Vision, Projector\}}
    \FOR{each targeted sub-module $W \in \mathcal{L}$}
        \STATE $\Delta W \leftarrow W^{\text{VLA}} - W^{\text{VLM}}$
        \STATE \textbf{Dimension Alignment:} Identify aggregation dim $d$
        \STATE \COMMENT{e.g., $d=1$ for input-side projection; $d=0$ for output-side projection}
        \STATE \textbf{Metric Calculation:} $v \leftarrow \|\Delta W\|_{L2, \text{dim}=d} \ / \ (\|W^{\text{VLM}}\|_{L2, \text{dim}=d} + \epsilon)$
        \IF{Target is LLM FFN}
            \STATE Accumulate $v$ across \{\textit{gate, up, down}\} projections to align FFN channels
        \ELSIF{Target is Attention Head}
            \STATE $v \leftarrow \text{Aggregate } v \text{ to head-level importance via mean operator}$
        \ENDIF
        \STATE Add aggregated scores $v$ to pool $S$
    \ENDFOR
\ENDFOR
\STATE \textbf{Global Ranking:} Find threshold $\tau$ for the \textbf{Lowest} or \textbf{Highest} $r\%$ elements in $S$
\STATE \textbf{Mask Assignment:} Generate unit mask $M_{\text{unit}}$ based on threshold $\tau$
\STATE \textbf{Intra-module Broadcast:} Expand $M_{\text{unit}}$ to parameter-wise masks $M$. 
\STATE \COMMENT{For FFNs, $M_{\text{unit}}$ masks the \textit{output} of fc1/up/gate and the \textit{input} of fc2/down simultaneously.}
\RETURN $M$
\end{algorithmic}
\end{algorithm}

\subsection{Unified Pipeline for Causal Analysis}
We employ a unified analysis pipeline: (1) Select correlated $\langle\text{VLM, VLA}\rangle$ pairs; (2) Calculate $\Delta W$ across sub-modules; (3) Analyze spatial distribution patterns; (4) Apply controlled pruning interventions; (5) Evaluate direct inference behavior without recovery; (6) Perform cross-module and cross-model comparisons. This methodology allows us to systematically identify which parameters are redundant and provides the theoretical support for exploration of recovery-free pruning strategies in \cref{sec:experiments}.

\section{Experiments}
\label{sec:experiments}

\subsection{Experimental Settings}
\textbf{Dataset.} We conduct experiments on the \textbf{LIBERO} benchmark \cite{liu2023libero}, the mainstream evaluation suite in embodied manipulation. LIBERO is designed to test skills inspired by human activities, requiring agents equipped with a Franka Panda arm to complete tasks based on natural language instructions and visual observations. The benchmark includes four sub-datasets: \textbf{LIBERO-Spatial} (same objects, different spatial layouts), \textbf{LIBERO-Object} (same layouts, different object categories), \textbf{LIBERO-Goal} (diverse task goals), and \textbf{LIBERO-Long} (long-horizon tasks). We use the Success Rate (SR) of each task set as the core performance metric.

\noindent \textbf{Baselines.} Our study is based on two representative VLM-to-VLA model pairs: $\langle\text{Prismatic, OpenVLA}\rangle$ and $\langle\text{PaLI-Gemma, } \pi_{0.5}\rangle$. The former represents the classic architecture adapted from large-scale LLMs (Llama-2 \cite{touvron2023llama}), while the latter represents emerging modular and lightweight VLA models. During the discovery and verification phases (\cref{sec:Hypothesis I}--\cref{sec:Hypothesis III}), we use these original models as Backbone VLA baselines. In the final algorithm application experiments (\cref{sec:Hypothesis IV}), we introduce widely-adopted pruning criteria as comparative baselines to evaluate their efficacy in a recovery-free setting, including structured pruning methods \textbf{LLM-Pruner}\cite{ma2023llm}, \textbf{FLAP}\cite{an2024fluctuation}, and the importance-based sparsification method \textbf{Wanda} \cite{sun2023simple}.

\noindent \textbf{Implementation Details.} All experiments are executed on NVIDIA A100 GPU platforms. In diagnostic experiments without performance recovery, models are evaluated via direct inference after pruning. In control experiments involving recovery (\cref{sec:Hypothesis I}), we utilize \textbf{LoRA} fine-tuning under the FSDP distributed strategy.

\subsection{Hypothesis I: If Pruning in VLA Models Requires Post-Pruning Performance Recovery, the Removed Parameters Are Unlikely to Be Truly Redundant}
\label{sec:Hypothesis I}
This subsection examines the "fine-tuning compensation paradigm" in VLA pruning. We question whether post-pruning performance recovery stems from true redundancy or the repair of "falsely killed" vital parameters. Using the Prismatic-OpenVLA pair, we compare three strategies—\textit{Highest-diff}, \textit{Lowest-diff}, and \textit{Random}—targeting the LLM's FFN intermediate layers. We evaluate performance pre- and post-LoRA fine-tuning (FSDP, 10k steps, LR 1e-4).

\begin{table}[t]
\centering
\caption{Model performance comparison before and after fine-tuning. LIBERO-Spatial, Baseline SR = 84.7\%.}
\label{tab:hypo1}
\footnotesize 
\setlength{\tabcolsep}{0pt} 
\begin{tabular*}{\linewidth}{@{\extracolsep{\fill}} l c cc}
\toprule
\textbf{Pruning Strategy} & \textbf{Ratio (\%)} & \textbf{SR (Pre-FT) (\%)} & \textbf{SR (Post-FT) (\%)} \\ \midrule
Lowest-diff  & 20 & 1.5  & 86.5 \\
Lowest-diff  & 50 & 0.0  & 81.0 \\
Lowest-diff  & 80 & 0.0  & 76.4 \\ \midrule
Highest-diff & 20 & 76.3 & 85.8 \\
Highest-diff & 50 & 20.5 & 84.1 \\
Highest-diff & 80 & 0.0  & 80.7 \\ \midrule
Random       & 20 & 12.2 & 86.0 \\
Random       & 50 & 0.0  & 82.2 \\
Random       & 80 & 0.0  & 77.6 \\
\bottomrule
\end{tabular*}
\end{table}

Table~\ref{tab:hypo1} reveals a critical paradox: in immediate evaluations of the LLM FFN, removing low-divergence channels (\textit{Lowest-diff}) causes total performance collapse. However, LoRA fine-tuning enables all configurations to recover to or exceed baseline levels, regardless of initial damage (even from 0.0\% SR). This "strong compensation" masks the inherent quality of pruning decisions. 

\begin{figure}[t]
    \centering
    \includegraphics[width=0.7\textwidth]{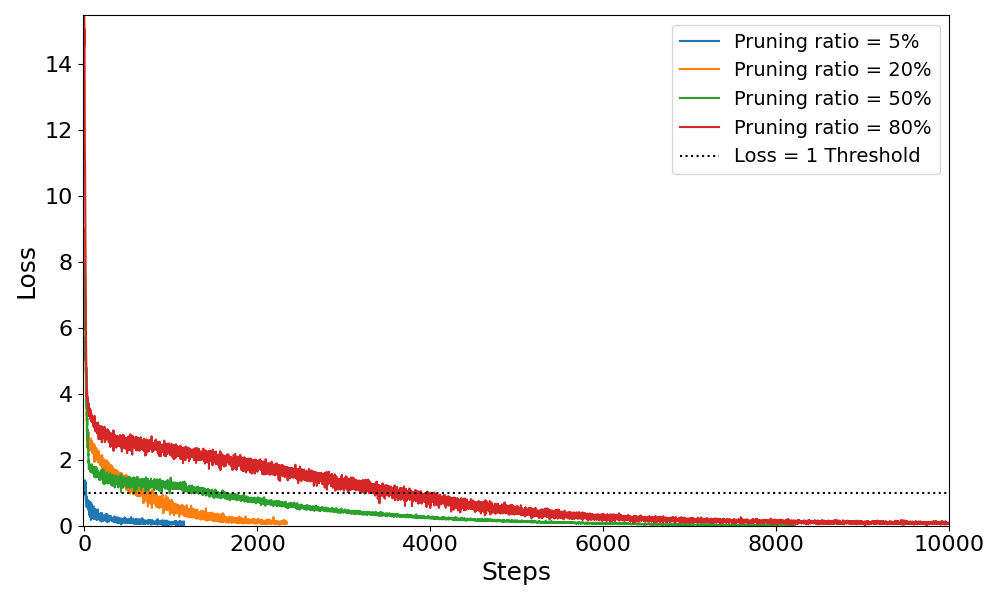}
    \caption{Causal analysis of recovery difficulty: Convergence steps vs. Pruning ratio.}
    \label{fig:steps_loss}
\end{figure}

Further causal analysis (Fig.~\ref{fig:steps_loss}) shows that convergence steps required for recovery scale with the pruning ratio. This increasing difficulty suggests that higher pruning ratios introduce deeper structural perturbations. These findings support \textbf{Hypothesis I}: reliance on recovery essentially repairs structural damage from "parameter mis-deletion." Thus, effective redundancy identification must maintain core functionality without requiring recovery.

\subsection{Hypothesis II: The Parameter Differences Introduced During VLM-to-VLA Adaptation Contain Useful Signals for Identifying Redundant Parameters}
\label{sec:Hypothesis II}
We systematically evaluate the effectiveness of $\Delta W$ in identifying redundancy by comparing \textit{Highest-diff} and \textit{Lowest-diff} pruning across OpenVLA and $\pi_{0.5}$. The results reveal that parameter divergence is not a global importance metric but exhibits profound \textbf{Module Heterogeneity}.

In OpenVLA's DINOv2 (Table~\ref{tab:tab2}), we observe a striking "sensitivity reversal": for attention heads, pruning the highest-diff heads causes collapse (1.6\% SR), whereas for FFN channels, pruning the lowest-diff channels leads to collapse (0.0\%). This indicates that functional importance is highly path-dependent even within the same backbone. Similar patterns emerge in the language: removing the lowest-diff attention heads or FFN channels leads to total degradation (0.0\% and 2.7\% SR, respectively), while pruning the results in total degradation (0.0\% and 2.7\% SR, respectively), whereas the highest-diff components maintain high performance. In contrast, SigLIP shows minimal sensitivity to $\Delta W$, consistent with its role as an auxiliary semantic supplement \cite{kim2024openvla}.

The modular $\pi_{0.5}$ model (Table~\ref{tab:tab3}) exhibits even clearer trends. Pruning highest-diff FFN channels in both vision and language barely affects performance ($\sim$95.0\% SR), while removing lowest-diff channels causes significant drops. Language model attention also collapses (0.0\%) only when the lowest-diff heads are removed. These cross-model observations confirm \textbf{Hypothesis II}: VLM-to-VLA parameter divergence contains structured signals that effectively distinguish vital from redundant parameters across different computational roles. 
\begin{table}[t]
\centering
\footnotesize
\setlength{\tabcolsep}{0pt}

\caption{Impact of VLM-VLA parameter differences across modules (OpenVLA). LIBERO-Spatial, Baseline SR = 84.7\%.}
\label{tab:tab2}

\begin{tabular*}{\linewidth}{@{\extracolsep{\fill}} llc cc}
\toprule
\multirow{2}{*}{\textbf{Module}} & \multirow{2}{*}{\textbf{Sub-module}} & \multirow{2}{*}{\textbf{Ratio}} & \multicolumn{2}{c}{\textbf{SR (\%)}} \\ 
\cmidrule{4-5}
& & & \textbf{High-diff (H)} & \textbf{Low-diff (L)} \\ 
\midrule
Vision & DINOv2 Attn (head) & 0.125 & 1.6 & 76.7 \\
Vision & DINOv2 FFN (channel) & 0.20 & 82.0 & 0.0 \\
Vision & SigLIP Attn (head) & 0.125 & 83.4 & 80.7 \\
Vision & SigLIP FFN (channel) & 0.20 & 75.1 & 81.7 \\
Language & Llama2 Attn (head) & 0.125 & 84.3 & 0.0 \\
Language & Llama2 FFN (channel) & 0.20 & 72.0 & 2.7 \\
Projector & FFN (channel) & 0.30 & 0.0 & 54.0 \\
\bottomrule
\end{tabular*}
\end{table}

\begin{table}[t]
\centering
\footnotesize
\setlength{\tabcolsep}{0pt}

\caption{Impact of VLM-VLA parameter differences across modules ($\pi_{0.5}$). LIBERO-Spatial, Baseline SR = 98.8\%.}
\label{tab:tab3}

\begin{tabular*}{\linewidth}{@{\extracolsep{\fill}} llc cc}
\toprule
\multirow{2}{*}{\textbf{Module}} & \multirow{2}{*}{\textbf{Sub-module}} & \multirow{2}{*}{\textbf{Ratio}} & \multicolumn{2}{c}{\textbf{SR (\%)}} \\ 
\cmidrule{4-5}
& & & \textbf{High-diff (H)} & \textbf{Low-diff (L)} \\ 
\midrule
Language & Gemma Attn (head) & 0.20 & 85.0 & 0.0 \\
Language & Gemma FFN (channel) & 0.50 & 95.0 & 5.0 \\
Vision & SigLIP Attn (head) & 0.20 & 90.0 & 55.0 \\
Vision & SigLIP FFN (channel) & 0.50 & 95.0 & 15.0 \\
\bottomrule
\end{tabular*}
\end{table}

\subsection{Hypothesis III: The Usefulness of VLM-VLA Parameter Differences for Redundancy Varies Across Modules}
\label{sec:Hypothesis III}
We further explore the boundaries of the $\Delta W$ signal utility. We find that the contribution of VLA components to the final performance is unequal, determining the strategic focus of global compression.

The Projector module is "fragile and non-selective." As shown in Table~\ref{tab:tab4}, across pruning ratios of 0.2 to 0.5, both pruning strategies lead to collapse. This suggests the Projector functions primarily as a cross-modal alignment interface; once this bottleneck is damaged, performance degrades regardless of which parameters are removed. It must be strictly protected. In contrast, SigLIP exhibits extreme robustness. Even with a pruning ratio of 1.0 (total removal), OpenVLA maintains moderate success. This contrasts with DINOv2, where even slight pruning causes failure. DINOv2 serves as the primary structural representation module, while SigLIP only provides supplementary semantic information. These results validate \textbf{Hypothesis III}: the signal's usability depends on the module's functional role, requiring differentiated allocation logic in joint pruning. 

\begin{table}[t]
\centering
\caption{Pruning robustness of weakly-coupled modules (OpenVLA). LIBERO-Spatial, Baseline SR = 84.7\%.}
\label{tab:tab4}
\footnotesize 
\setlength{\tabcolsep}{0pt}
\begin{tabular*}{\linewidth}{@{\extracolsep{\fill}} llc cc}
\toprule
\multirow{2}{*}{\textbf{Module}} & \multirow{2}{*}{\textbf{Sub-module}} & \multirow{2}{*}{\textbf{Ratio}} & \multicolumn{2}{c}{\textbf{SR (\%)}} \\ \cmidrule{4-5}
& & & \textbf{High-diff (H)} & \textbf{Low-diff (L)} \\ \midrule
Projector & FFN (channel) & 0.20 & 84.5 & 78.5 \\
Projector & FFN (channel) & 0.30 & 0.0 & 54.0 \\
Projector & FFN (channel) & 0.50 & 0.0 & 0.0 \\ \midrule
Vision & SigLIP Attn (head) & 1.00 & 47.0 & -- \\
Vision & SigLIP FFN (channel) & 1.00 & 70.0 & -- \\
\bottomrule
\end{tabular*}
\end{table}

\subsection{Hypothesis IV: When Properly Utilized, VLM-VLA Parameter Differences Allow Structured Pruning Without Requiring Post-Pruning Performance Recovery}
\label{sec:Hypothesis IV}
We verify if the revealed module heterogeneity enables a \textbf{recovery-free } pruning scheme. We design a multi-module joint pruning algorithm with three configurations: \textit{Pruned-Light, Moderate,} and \textit{Aggressive}. Following \cite{chen2025rlrc,jabbour2025don}, we compare our approach against representative pruning criteria evaluated under identical recovery-free conditions: \textbf{LLM-Pruner}~\cite{ma2023llm}, \textbf{FLAP}~\cite{an2024fluctuation}, and \textbf{Wanda}~\cite{sun2023simple}.

Table~\ref{tab:tab5} shows results on OpenVLA. Traditional methods (LLM-Pruner, FLAP, Wanda) suffer catastrophic failure in the absence of performance recovery. In a \textbf{matched-capacity comparison} at 12.4GB, our \textit{Moderate} config maintains 62.3\% SR, while LLM-Pruner achieves only 1.0\%. This 60\% gap demonstrates that conventional pruning metrics fail in VLA because it removes vital parameters undergoing functional reorganization. In the \textit{Aggressive} config (5.7B), the model still retains 60\% efficacy in core tasks like Spatial and Object.

The pattern holds for $\pi_{0.5}$ (Table~\ref{tab:tab6}). Our \textit{Moderate} scheme reduces memory from 7.3GB to 5.6GB, with SR only shifting from 96.9\% to 89.0\%. This supports \textbf{Hypothesis IV}: VLA pruning sensitivity is not insurmountable. By correctly interpreting structural importance from adaptation, we can achieve efficient deployment on edge devices without expensive compensation.

\begin{table}[t]
\centering
\caption{Task success rates (\%) on LIBERO benchmark for pruned OpenVLA variants.}
\label{tab:tab5}
\renewcommand{\arraystretch}{1.0}
\begin{tabularx}{\linewidth}{Xccccccc}
\toprule
\textbf{Model} & \textbf{Params} & \textbf{Mem} & \textbf{Spatial} & \textbf{Object} & \textbf{Goal} & \textbf{Long} & \textbf{Average} \\
 & \textbf{(B)} & \textbf{(GB)} & \textbf{(\%)} & \textbf{(\%)} & \textbf{(\%)} & \textbf{(\%)} & \textbf{(\%)} \\
\midrule
OpenVLA (Baseline) & 7.5 & 14.9 & 84.7 & 88.4 & 79.2 & 53.7 & 76.5 (+0.0) \\
\midrule
LLM-Pruner \cite{chen2025rlrc} & 6.2 & 12.4 & 23.4 & - & - & 1.0 & - \\
FLAP \cite{chen2025rlrc} & 6.3 & 12.5 & 0.2 & - & - & 0.0 & - \\
Wanda(Full Sparse) \cite{jabbour2025don} & - & 10.2 & 0.0 & 13.4 & 0.8 & 0.0 & 7.1 (-69.4) \\
Wanda(Sparse Lang. BB) \cite{jabbour2025don} & - & 10.6 & 31.2 & 50.8 & 20.0 & 12.4 & 28.6 (-47.9) \\
Wanda(75\% Sparse Lang. BB) \cite{jabbour2025don} & - & 12.0 & 25.4 & 49.0 & 20.8 & 11.4 & 26.7 (-49.8) \\
\midrule
\textbf{Ours-Light} & 6.6 & 13.0 & 78.3 & 82.5 & 74.0 & 46.8 & \textbf{70.4 (-6.1)} \\
\textbf{Ours-Moderate} & 6.2 & 12.4 & 70.5 & 74.9 & 64.7 & 39.0 & \textbf{62.3 (-14.2)} \\
\textbf{Ours-Aggressive} & 5.7 & 11.3 & 59.0 & 65.7 & 56.0 & 29.5 & \textbf{52.5 (-24.0)} \\
\bottomrule
\end{tabularx}
\end{table}

\begin{table}[t]
\centering
\caption{Task success rates (\%) on LIBERO benchmark for pruned $\pi_{0.5}$ variants.}
\label{tab:tab6}
\renewcommand{\arraystretch}{1.05}
\begin{tabularx}{\linewidth}{Xccccccc}
\toprule
\textbf{Model} & \textbf{Params} & \textbf{Mem} & \textbf{Spatial} & \textbf{Object} & \textbf{Goal} & \textbf{Long} & \textbf{Average} \\
 & \textbf{(B)} & \textbf{(GB)} & \textbf{(\%)} & \textbf{(\%)} & \textbf{(\%)} & \textbf{(\%)} & \textbf{(\%)} \\
\midrule
$\pi_{0.5}$ (Baseline) & 3.6 & 7.3 & 98.8 & 98.2 & 98.0 & 92.4 & 96.9 (+0.0) \\
\midrule
\textbf{Ours-Light} & 3.0 & 6.1 & 95.5 & 94.0 & 95.0 & 88.5 & \textbf{93.3 (-3.6)} \\
\textbf{Ours-Moderate} & 2.8 & 5.6 & 90.7 & 90.1 & 91.3 & 84.0 & \textbf{89.0 (-7.9)} \\
\textbf{Ours-Aggressive} & 2.5 & 5.0 & 83.6 & 83.0 & 84.5 & 77.1 & \textbf{82.1 (-14.8)} \\
\bottomrule
\end{tabularx}
\end{table}

\section{Conclusion}
This study systematically investigates parameter redundancy and parameter pruning sensitivity in VLA models by focusing on the structured signals inherent in VLM-to-VLA adaptation. Our analysis reveals that the prevailing reliance on post-hoc recovery often masks the "indiscriminate killing" of vital parameters, whereas adaptation-induced parameter divergence ($\Delta W$) serves as a high-fidelity criterion for parameter redundancy identification. We uncover significant structural heterogeneity across modules: primary structural vision encoders and language backbones exhibit high pruning sensitivity, cross-modal projectors act as fragile alignment bottlenecks, while semantic-supplementary vision towers demonstrate remarkable redundancy resilience. By leveraging these insights through differentiated pruning logic, we demonstrate across multiple representative VLA baselines that substantial model compression is achievable without any fine-tuning or compensation mechanisms. Ultimately, understanding parameter evolution from general perception to embodied action is crucial for efficient VLA deployment, providing a foundation for future research in resource-constrained robotic intelligence.

\section*{Acknowledgements}
This work was supported by the Strategic Priority Research Program of Chinese Academy of Sciences under Grant No. XDA0460301 and the National Natural Science Foundation of China under Grant No. 62506235.

%
%
\bibliographystyle{plain}
\bibliography{main}

\clearpage
\appendix

\section{Detailed Model Composition and Parameter Dimensions}
\label{sec:appendix_composition}

In this section, we provide a comprehensive breakdown of the parameter composition for the two primary VLA models utilized in our study: OpenVLA (based on the Prismatic VLM) and $\pi_{0.5}$ (based on PaliGemma). Our pruning analysis specifically targets the weight parameters within the Vision Tower, Projector, and Language Model modules.

\subsection{Parameter Module Distribution}
To illustrate the structural complexity of the VLA baselines, we first count the number of individual weight tensors (including biases and normalization parameters) across the functional modules. As shown in Table~\ref{tab:parameter_count}, the majority of the parameter objects are concentrated in the Vision Backbone and Language Model.

\begin{table}[h]
\centering
\caption{Model parameter specifications and module distribution.} 
\label{tab:parameter_count}
\footnotesize 
\setlength{\tabcolsep}{0pt}
\begin{tabular*}{\textwidth}{@{\extracolsep{\fill}} l cccccc }
\toprule
\textbf{Model} & \textbf{Total} & \textbf{Vision} & \textbf{Projector} & \textbf{Language} & \textbf{Expert} & \textbf{Other} \\ \midrule
OpenVLA        & 982            & 685             & 6                  & 291               & --              & --             \\
$\pi_{0.5}$    & 812            & 437             & 2                  & 164               & 201             & 8              \\
\bottomrule
\end{tabular*}
\end{table}

\subsection{Detailed Weight Dimensions for Pruning}
Our pruning methodology focuses on the fundamental channel units within the Attention and Feed-Forward Network (FFN) blocks. Table~\ref{tab:weight_dims} details the dimensions for the key weight-bearing layers. Note that for the Language Model in $\pi_{0.5}$, Multi-Query Attention (MQA) is employed, resulting in fewer $K/V$ heads relative to $Q$ heads.

\begin{table}[h]
\centering
\caption{Architectural dimensions for the primary weight-bearing modules. $d_{model}$: hidden size; $L$: number of layers; $N_{head}$: attention heads; $d_{int}$: intermediate size.}
\label{tab:weight_dims}
\footnotesize 
\setlength{\tabcolsep}{0pt}
\begin{tabular*}{\textwidth}{@{\extracolsep{\fill}} ll cccc }
\toprule
\textbf{Model Family} & \textbf{Component} & \textbf{$d_{model}$} & \textbf{$L$} & \textbf{$N_{head}$} & \textbf{$d_{int}$} \\ \midrule
\multirow{3}{*}{OpenVLA} & Vision (DINOv2)     & 1024                & 24           & 16                  & 4096              \\
                         & Vision (SigLIP)     & 1152                & 27           & 16                  & 4304              \\
                         & Language (Llama-2)  & 4096                & 32           & 32                  & 11008             \\ \midrule
\multirow{2}{*}{$\pi_{0.5}$}   & Vision (SigLIP)  & 1152                & 27           & 16                  & 4304              \\
                         & Language (Gemma)    & 2048                & 18           & 8                   & 16384             \\ \bottomrule
\end{tabular*}
\vspace{1mm}
\begin{flushleft}
\scriptsize{$^\ast$ In $\pi_{0.5}$, $Q$ uses 8 heads while $K/V$ use 1 head (MQA) to reduce memory overhead during inference.}
\end{flushleft}
\end{table}

\subsection{Projector Configurations}
The cross-modal projectors serve as the bridge between vision and language.
\begin{itemize}
    \item \textbf{OpenVLA:} Uses a 2-layer MLP (3 weight tensors) with dimensions mapping from the fused vision tokens ($1024+1152=2176$) to the Llama hidden size ($4096$).
    \item \textbf{$\pi_{0.5}$:} Employs a single linear projection (2 weight tensors including bias) mapping from $1152$ to $2048$.
\end{itemize}

\section{Extended Experimental Results}
\label{sec:appendix_more_data}

This section provides a complete empirical verification of our hypotheses across all LIBERO sub-datasets. We provide granular data for both OpenVLA and $\pi_{0.5}$ to demonstrate the universality of our observations.

\subsection{Generalizability of Hypothesis I: The Recovery Paradox}
Table~\ref{tab:appendix_hypo1_all} illustrates the "strong compensation" effect. Across all task suites and both model families, fine-tuning consistently bridges the performance gap caused by sub-optimal pruning decisions, reinforcing the need for recovery-free evaluation.

\begin{table}[h]
\centering
\caption{Pre- and Post-finetuning Success Rate (SR \%) across all LIBERO sub-datasets. Pruning targets: Llama-2 FFN (20\%) for OpenVLA. \textit{Pre.} and \textit{Post.} denote results before and after fine-tuning, respectively.}
\label{tab:appendix_hypo1_all}
\footnotesize 
\setlength{\tabcolsep}{0pt} 
\begin{tabular*}{\textwidth}{@{\extracolsep{\fill}} ll cc cc cc cc }
\toprule
\multirow{2}{*}{\textbf{Model}} & \multirow{2}{*}{\textbf{Strategy}} & \multicolumn{2}{c}{\textbf{Spatial}} & \multicolumn{2}{c}{\textbf{Object}} & \multicolumn{2}{c}{\textbf{Goal}} & \multicolumn{2}{c}{\textbf{Long}} \\ 
\cmidrule{3-4} \cmidrule{5-6} \cmidrule{7-8} \cmidrule{9-10}
 &  & \textit{Pre.} & \textit{Post.} & \textit{Pre.} & \textit{Post.} & \textit{Pre.} & \textit{Post.} & \textit{Pre.} & \textit{Post.} \\ \midrule
\multirow{4}{*}{OpenVLA} & Baseline & \multicolumn{2}{c}{84.7} & \multicolumn{2}{c}{88.4} & \multicolumn{2}{c}{79.2} & \multicolumn{2}{c}{53.7} \\ 
\cmidrule{2-10}
 & Lowest-diff & 1.5 & 86.5 & 3.2 & 86.8 & 0.0 & 79.5 & 0.0 & 49.5 \\
 & Highest-diff & 76.3 & 85.8 & 78.4 & 89.5 & 71.2 & 81.6 & 44.8 & 53.9 \\
 & Random & 12.2 & 86.0 & 14.8 & 87.2 & 7.6 & 76.8 & 4.2 & 51.2 \\ \bottomrule
\end{tabular*}
\end{table}

\subsection{Validation of Hypothesis II: Signal Validity and Module Heterogeneity}
We evaluate the contrastive impact of \textit{High-diff} (H) and \textit{Low-diff} (L) strategies. As shown in Table~\ref{tab:appendix_hypo2_contrast}, the "sensitivity reversal" (e.g., in DINOv2 vs. Llama-2) is a persistent structural property. We have included SigLIP and Projector data for OpenVLA to provide a full-spectrum analysis of the adaptation signal.

\begin{table}[h]
\centering
\caption{Contrastive analysis of \textbf{High-diff (H)} vs. \textbf{Low-diff (L)} strategies (SR \%) across multiple modules. Note the performance fluctuations reflecting inherent task difficulty in the Long-horizon suite.}
\label{tab:appendix_hypo2_contrast}
\footnotesize 
\setlength{\tabcolsep}{0pt}
\begin{tabular*}{\textwidth}{@{\extracolsep{\fill}} llc cc cc cc cc }
\toprule
\multirow{2}{*}{\textbf{Model}} & \multirow{2}{*}{\textbf{Module}} & \multirow{2}{*}{\textbf{Ratio}} & \multicolumn{2}{c}{\textbf{Spatial}} & \multicolumn{2}{c}{\textbf{Object}} & \multicolumn{2}{c}{\textbf{Goal}} & \multicolumn{2}{c}{\textbf{Long}} \\ \cmidrule{4-5} \cmidrule{6-7} \cmidrule{8-9} \cmidrule{10-11}
 &  &  & \textbf{H} & \textbf{L} & \textbf{H} & \textbf{L} & \textbf{H} & \textbf{L} & \textbf{H} & \textbf{L} \\ \midrule
\multirow{6}{*}{OpenVLA} & DINOv2 Attn & 0.125 & 1.6 & \textbf{76.7} & 5.4 & \textbf{79.8} & 2.1 & \textbf{73.5} & 0.0 & \textbf{44.2} \\
 & DINOv2 FFN & 0.20 & \textbf{82.0} & 0.0 & \textbf{83.7} & 3.2 & \textbf{75.4} & 1.2 & \textbf{50.6} & 0.0 \\
 & SigLIP Attn & 0.125 & 83.4 & 79.7 & 85.2 & 81.5 & 77.1 & 75.2 & 48.6 & 42.0 \\
 & SigLIP FFN & 0.20 & 75.1 & 81.0 & 76.0 & 82.2 & 65.5 & 70.1 & 43.5 & 48.0 \\
 & Llama2 Attn & 0.125 & \textbf{84.3} & 0.0 & \textbf{87.1} & 0.0 & \textbf{78.4} & 0.0 & \textbf{52.5} & 0.0 \\
 & Llama2 FFN & 0.20 & \textbf{72.0} & 2.7 & \textbf{79.2} & 4.8 & \textbf{66.5} & 2.4 & \textbf{41.8} & 0.8 \\ \midrule
\multirow{4}{*}{$\pi_{0.5}$} & SigLIP Attn & 0.20 & \textbf{90.0} & 55.0 & \textbf{90.0} & 60.0 & \textbf{85.0} & 50.0 & \textbf{80.0} & 45.0 \\
 & SigLIP FFN & 0.50 & \textbf{95.0} & 15.0 & \textbf{90.0} & 20.0 & \textbf{90.0} & 15.0 & \textbf{85.0} & 10.0 \\
 & Gemma Attn & 0.20 & \textbf{85.0} & 0.0 & \textbf{85.0} & 5.0 & \textbf{80.0} & 5.0 & \textbf{75.0} & 0.0 \\
 & Gemma FFN & 0.50 & \textbf{95.0} & 5.0 & \textbf{95.0} & 10.0 & \textbf{90.0} & 5.0 & \textbf{85.0} & 5.0 \\ \bottomrule
\end{tabular*}
\end{table}

\subsection{Validation of Hypothesis III: Signal Boundaries and Robustness}
This subsection explores the limits of pruning robustness. Table~\ref{tab:appendix_hypo3_robust} demonstrates the extreme robustness of SigLIP and the severe fragility of the Projector across all task suites. The Projector's collapse under both strategies at Ratio 0.3 identifies it as a critical non-selective bottleneck.

\begin{table}[h]
\centering
\caption{Robustness limits of weakly-coupled and bottleneck modules (SR \%). Ratio 1.0 represents the complete removal of the component.}
\label{tab:appendix_hypo3_robust}
\footnotesize 
\setlength{\tabcolsep}{0pt}
\begin{tabular*}{\textwidth}{@{\extracolsep{\fill}} llc cc cc cc cc }
\toprule
\multirow{2}{*}{\textbf{Model}} & \multirow{2}{*}{\textbf{Module}} & \multirow{2}{*}{\textbf{Ratio}} & \multicolumn{2}{c}{\textbf{Spatial}} & \multicolumn{2}{c}{\textbf{Object}} & \multicolumn{2}{c}{\textbf{Goal}} & \multicolumn{2}{c}{\textbf{Long}} \\ \cmidrule{4-5} \cmidrule{6-7} \cmidrule{8-9} \cmidrule{10-11}
 &  &  & \textbf{H} & \textbf{L} & \textbf{H} & \textbf{L} & \textbf{H} & \textbf{L} & \textbf{H} & \textbf{L} \\ \midrule
\multirow{3}{*}{OpenVLA} & SigLIP Attn & 1.00 & 47.0 & -- & 51.5 & -- & 40.8 & -- & 27.2 & -- \\
 & SigLIP FFN & 1.00 & 70.0 & -- & 73.8 & -- & 64.2 & -- & 38.5 & -- \\
 & Projector FFN & 0.30 & 0.0 & \textbf{54.0} & 0.0 & \textbf{57.4} & 0.0 & \textbf{50.8} & 0.0 & \textbf{26.5} \\ \bottomrule
\end{tabular*}
\end{table}

\section{Algorithmic Specifications for Joint Pruning}
To ensure reproducibility, we provide the detailed configurations for the \textit{Light, Moderate,} and \textit{Aggressive} settings used in our multi-module joint pruning scheme. As shown in Table~\ref{tab:config}, different modules are assigned specific pruning ratios and selection criteria based on the divergence signals $\Delta W$ observed during VLM-to-VLA adaptation.

\begin{table}[h]
\centering
\caption{Pruning configurations for OpenVLA and $\pi_{0.5}$. Ratios indicate the fraction of parameters removed. Selection criteria: \textbf{H} (High-diff), \textbf{L} (Low-dif).}
\label{tab:config}
\footnotesize
\renewcommand{\arraystretch}{1.1}
\begin{tabularx}{\linewidth}{l|ccc|ccc}
\toprule
 & \multicolumn{3}{c|}{\textbf{OpenVLA}} & \multicolumn{3}{c}{\textbf{$\pi_{0.5}$}} \\
\textbf{Module} & \textbf{Light} & \textbf{Moderate} & \textbf{Aggressive} & \textbf{Light} & \textbf{Moderate} & \textbf{Aggressive} \\
\midrule
LLM-Attn & 0.125 (H) & 0.125 (H) & 0.125 (H) & 0.2 (H) & 0.2 (H) & 0.2 (H) \\
LLM-FFN & 0.1 (H) & 0.2 (H) & 0.3 (H) & 0.2 (H) & 0.3 (H) & 0.5 (H) \\
SigLIP-Attn & 0.125 (H) & 0.125 (H) & 0.125 (H) & 0.2 (H) & 0.2 (H) & 0.2 (H) \\
SigLIP-FFN & 0.1 (L) & 0.1 (L) & 0.1 (L) & 0.4 (H) & 0.2 (H) & 0.2 (H) \\
DINOv2-Attn & 0.0625 (L) & 0.0625 (L) & 0.0625 (L) & - & - & - \\
DINOv2-FFN & 0.1 (H) & 0.1 (H) & 0.1 (H) & - & - & - \\
\bottomrule
\end{tabularx}
\end{table}

\section{Generalization Across Models and Benchmarks}
\label{sec:appendix_generalization}

To further validate the generality of our adaptation-based pruning criterion, we evaluate $\pi_0$ on five RoboTwin2.0 tasks by pruning the highest- and lowest-$\Delta W$ 10\% channels in the LLM FFN layers.

As shown in Table~\ref{tab:appendix_pi0_robotwin}, pruning high-$\Delta W$ channels consistently outperforms pruning low-$\Delta W$ channels across all tasks, demonstrating that adaptation-induced parameter divergence remains an effective indicator of parameter importance across different VLA models and benchmarks.

\begin{table}[h]
\centering
\caption{
Generalization evaluation on RoboTwin2.0 using $\pi_0$.
We compare pruning the highest- and lowest-$\Delta W$ 10\% channels in the LLM FFN layers.
}
\label{tab:appendix_pi0_robotwin}

\footnotesize
\setlength{\tabcolsep}{5pt}
\renewcommand{\arraystretch}{1.1}

\begin{tabular}{lccc}
\toprule
\textbf{Simulation Task}
&
\textbf{Original}
&
\textbf{High-$\Delta W$}
&
\textbf{Low-$\Delta W$}
\\

\midrule

Beat Block Hammer
&
41
&
35
&
8

\\

Move Can Pot
&
55
&
43
&
11

\\

Shake Bottle
&
90
&
80
&
18

\\

Place Phone Stand
&
32
&
27
&
6

\\

Rotate QRcode
&
62
&
52
&
10

\\

\midrule

\textbf{Average}
&
\textbf{56.0}
&
\textbf{47.4}
&
\textbf{10.6}

\\

\bottomrule
\end{tabular}

\end{table}

\end{document}